\documentclass{article}

\usepackage[preprint]{neurips_2025}
\workshoptitle{}

\usepackage{amsmath,amssymb,amsfonts}
\usepackage{booktabs}
\usepackage{graphicx}
\usepackage{wrapfig}
\usepackage{url}
\usepackage[hidelinks]{hyperref}
\usepackage{xcolor}
\usepackage{enumitem}
\usepackage{caption}

\usepackage{pgfplots}
\pgfplotsset{compat=1.18}
\usepackage{tcolorbox}
\usepackage{listings}
\usepackage{float}
\definecolor{codegray}{gray}{0.95}
\definecolor{keywordcolor}{rgb}{0.0,0.0,0.6}
\definecolor{stringcolor}{rgb}{0.75,0.1,0.0}
\definecolor{commentcolor}{rgb}{0.0,0.5,0.0}
\definecolor{identifiercolor}{rgb}{0.0,0.0,0.0}
\lstdefinestyle{arxivpython}{
    language=Python,
    backgroundcolor=\color{codegray},
    basicstyle=\ttfamily\small,
    keywordstyle=\color{keywordcolor}\bfseries,
    stringstyle=\color{stringcolor},
    commentstyle=\color{commentcolor}\itshape,
    identifierstyle=\color{identifiercolor},
    breaklines=true,
    showstringspaces=false,
    tabsize=4,
    morekeywords={f}, columns=fullflexible,
}

\title{Multilingual VLM Training: Adapting an English-Trained VLM to French}

\author{Jules Lahmi\textsuperscript{\rm 1} \qquad Alexis Roger\textsuperscript{\rm 2,3} \\
\textsuperscript{\rm 1}Ecole Polytechnique,
\textsuperscript{\rm 2}Mila - Quebec AI Institute, 
\textsuperscript{\rm 3}McGill University
}

\begin{document}
\maketitle

\begin{abstract}
Artificial intelligence has made great progress in recent years, particularly in the development of Vision--Language Models (VLMs) that understand both visual and textual data. However, these advancements remain largely limited to English, reducing their accessibility for non--English speakers. It is essential to extend these capabilities to a broader range of languages. This paper explores the challenges of adapting an English-trained VLM to different languages. To this end, we will explore and compare different methods for their performance and computational cost. We consider
a translation-based pipeline, LoRA finetuning, and a two-stage finetuning strategy that separates vision adaptation from language adaptation. To evaluate these methods, we use a combination of standard multimodal benchmarks translated into the target language and manual assessments by native experts. The results reveal that dataset translation remains a major bottleneck in multilingual VLM performance, with data quality limiting the effectiveness of training and evaluation. These findings suggest that future efforts should focus on native-language dataset collection and improved translation strategies.

\end{abstract}

\section{Introduction}
With Vision-Language Models (VLMs) emerging as powerful tools 
capable of interpreting and generating text based on visual input ~\citep{VLM},
artificial intelligence is rapidly evolving and being used in multiple fields, making them increasingly relevant in both research and industry. 

A lot of high-quality VLMs have been trained in english 
using popular language models such as Llama ~\citep{Llama3} or Mistral ~\citep{Mistral7B}, but the ability to train VLMs in multiple languages expands their accessibility.
Therefore, we will focus our efforts on adapting an English-trained VLM new languages as a first step toward broader 
multilingual VLM development.
In this paper, we will be using French as a target language and proof of concept, but this methods hold for any target language chosen. We explore different strategies 
for adapting the model, analyzing effectiveness, limitations and cost for generalization.

We aim to adapt an English VLM to operate in French without full retraining. The core difficulty we tackle is the scarcity of large, high-quality, French multimodal datasets and the limited availability of affordable translation at scale. We address this gap by translating existing English datasets into French for training and evaluation, providing a substitute for the absence of large French resources.

As translation affects pretraining, finetuning, and evaluation, it is important for us to obtain a high-quality translator. Cost constraints prevented use of commercial APIs such as DeepL or Google Translate. Nevertheless, we built a working translation pipeline used both as a baseline and as a data-generation tool.

Through evaluating translation choices, adaptation strategies, and prompt design, we aim to provide practical guidance for enabling multilingual capability in VLMs. Our main contributions are summarized as follows:
\begin{enumerate}\itemsep0pt
    \item We build a practical end-to-end translation pipeline (French $\longleftrightarrow{}$English) around an English-trained VLM, including highly effective prompts.
    \item We implement and compare two parameter-efficient adaptation strategies for adapting an English model to a target language (here French).
    \item We introduce a translation-quality audit that quantifies noise and yields actionable recommendations (e.g., when to prioritize native-language data; when data is scarce, I/O translation can outperform finetuning).
\end{enumerate}

\section{Methods}
\label{sec:methods}
For this study, we combined two high-performance models for vision and language processing, pairing the \textbf{OpenHermes~2.5 Mistral~7B LLM} ~\citep{Mistral7B}, with a \textbf{ViT SO400M 14 SigLIP 384} vision encoder  ~\citep{VIT} ~\citep{BigVision}.These are combined using a Llava architecture \citep{llava}. All translations in this work was done using \textbf{Llama-3.1-8B-Instruct}~\citep{Llama3}.  
In this work we will be using french as our target language, thanks to the easy availability of native speakers as well as the high use of the language. We recall that this is only for illustration and these methods can be used with any language.

The translation process required careful prompt design to ensure translation quality. Without strong instructions, the translator occasionally answered questions instead of translating, or removed the interrogative form of questions. Long, explicit prompts proved most effective in reducing this drift and allowed the system to operate consistently across datasets (see Appendix~\ref{app:prompt-translation}). 

We will compare the following 4 methods: a translation Pipeline with no finetuning (\textbf{1}), an English Pretraining with French Finetuning (\textbf{2}), a French Pretraining with French Finetuning (\textbf{3}), and a double Finetuning strategy (\textbf{4}).

\textbf{1. Translation pipeline with no finetuning}:
Our first method does not involve any training to see if current models can be used immediately. We translate each French prompt into English, process it with the English-trained VLM, and translate the output back into French (See Appendix~\ref{app:pipeline}). This baseline preserves the original model's capabilities by handling translation only at the input and output stages.

\textbf{2. English pretraining and French finetuning}:
In this method, we pretrain the VLM in English using standard methods. then finetune the VLM on a smaller French dataset. LoRA is used as it yields parameter efficiency with comparable quality to full finetuning in many settings~\citep{LoRA}.

For pretraining we use a sufficiently large subset of \texttt{Train\_Monkey\_2}~\citep{MonkeyDataset} to allow the loss curve to converge. For LoRA finetuning we use a translated subset of \texttt{blip\_laion\_cc\_sbu\_558k} (simple ``what is in this image?'' descriptions). Hyperparameters for both pretraining and LoRa finetuning were stable   (Appendix~\ref{appendix:Hyperparameters_finetuning}).

\textbf{3. French pretraining and French finetuning}:
We also create an alternate model where the pretraining was done in French using a translation of \texttt{Train\_Monkey\_2}. This model was trained using the same datasets as the previous one.

\textbf{4. Double finetuning in the target language}:
After pretraining in English, we first adapt the vision encoder on French data while effectively freezing the LLM, then LoRA-finetune the LLM on French with the vision encoder frozen.

To train this way, we first set the LLM learning rate to $1\times10^{-10}$ in stage~1, keeping the encoder LR high. Then in stage~2, we set the encoder LR was near-zero while LoRA-finetuning the LLM. We hope that this approach allows for better multimodal alignment while keeping reproducibility costs low.

\definecolor{mygreen}{RGB}{102, 205, 102} \definecolor{mydarkgreen}{RGB}{0,128,0}

\begin{figure}[b]
  \centering
\begin{minipage}[c]{0.48\linewidth}
    \centering
    \begin{tikzpicture}[scale=0.7]
      \begin{axis}[
          ybar,
          bar width=13pt, symbolic x coords={High, Moderate, Low},
          xtick=data,
          ymin=0, ymax=70,
          ylabel={Percentage of samples},
legend pos=north east,
          nodes near coords,
          enlarge x limits=0.2,
          x tick label style={align=center,font=\small},
      ]
      \addplot coordinates {(High, 62) (Moderate, 34) (Low, 4)};
      \addplot coordinates {(High, 60) (Moderate, 30) (Low, 10)};
      \legend{Questions, Answers}
      \end{axis}
    \end{tikzpicture}
    \caption{Histogram of the translation qualities}
    \label{fig:translation-quality-plot}
  \end{minipage}\hfill
\begin{minipage}[c]{0.48\linewidth}
    \centering
    \small \captionsetup{type=table}
    \caption{Joint question/answer translation quality matrix (percent of pairs).}
    \begin{tabular}{lccc}
      \toprule
      \textbf{Q$\backslash$A quality} & \textbf{Low} & \textbf{Moderate} & \textbf{High} \\
      \midrule
      \textbf{Low}      & {\color{red}0\%} & {\color{red}4\%} & {\color{red}0\%}  \\
      \textbf{Moderate} & {\color{red}2\%} & {\color{mygreen}12\%} & {\color{mygreen}20\%} \\
      \textbf{High}     & {\color{red}8\%} & {\color{mygreen}14\%} &{\color{mydarkgreen}40\%} \\
      \bottomrule
    \end{tabular}
    \label{tab:quality-matrix}
  \end{minipage}
\end{figure}

\section{Evaluation}
\label{sec:evaluation}

\subsection{Translation quality}
\label{sec:translation-quality}

We evaluate translation quality by comparing the original English dataset to its double-translated equivalent, where the target dataset is translated to the target language and back to english. In order to accurately evaluate the translation and grade them without a simple pass/fail, we will be evaluating translation manually, categorizing each translated question and answer in groups, and analyzing the
results. This will allow us to estimate our error margin on both the evaluation
and training datasets.

We manually categorize 200 Questions and their 200 Answers selected from \texttt{Train\_Monkey\_2} as:
\begin{enumerate}
\item \textbf{High}: both the essence and details were captured by the translation
    \item \textbf{Moderate}: the core meaning was preserved but interrogation or details were lost (See Appendix ~\ref{appendix:Lost_Interrogation})
    \item \textbf{Low}: there were errors or the model tried to answer the question (See Appendix ~\ref{appendix:low})
\end{enumerate}

\subsection{Benchmarks and scoring}
\textbf{General VQA} tasks necessitate the model ability to understand and integrate visual and textual information. This
involves a comprehensive grasp of how these modalities
interrelate. We validate our model using the 
ScienceQA \citep{ScienceQA} and POPE \citep{POPE} benchmarks. These benchmarks provide a broad assessment of
the model capabilities in general visual question answering
scenarios.

\textbf{Scene Text-centric VQA} tasks consist of text within images, as it is prevalent in real-world environments. For evaluating our model’s performance in this area, we utilized the
TextVQA \citep{TextVQA} benchmark.
\section{Results}
\label{sec:results}

\subsection{Translation model assessment}
Inspection of the round-trip translations shows a substantial margin of error, especially when the translator slips into answering mode or drops interrogatives. High-quality translation occurs in 60\% of translations, as shown in Figure~\ref{fig:translation-quality-plot}. Although suboptimal, these results are decent for a light model that can be run locally, or even using the same LLM as the VLM. These solutions allow for a more practical and cost-effective approach, compensating for the translation noise.

The main challenge to overcome is the association of a poorly translated question to a correct answer, or a poorly translated answer to a correct question. Indeed, as we can see in Table \ref{tab:quality-matrix}, the percentage of ’high quality’ question-answer pairs is only 40\%.
Since pairs with mixed quality (e.g., high-quality question, moderate-quality answer) degrade training and evaluation reliability, we estimate 60\% of pairs are unsuitable for precise evaluation or training.

\subsection{Loss curves}
For the English pretrained model (\textbf{2}), the French finetuning loss drops from $\sim$9\% to $\sim$2.0--2.5\% and stabilizes (Appendix \ref{app:loss-english-finetune}). This shows the model is well adapted to new data, without showing signs of overfitting, following expectations for a large-scale finetuning. For French pretrained and French finetuned model, the loss drops from $\sim$3\% to $\sim$1.8--2.2\% (Appendix \ref{app:loss-french-finetune}). This indicates better initialization for French tasks since the model has already been exposed to French in the pretraining. We also notice the final loss is lower that on the English pretrained model, so we expect this model to perform better. The models are thus effectively learning on the training data at every step of the training.

\subsection{Benchmark accuracy}
\begin{table}[t]
  \centering
  \caption{Accuracy (\%) across benchmarks. Asterisks denote manual evaluations.}
  \begin{tabular}{lcccccc}
    \toprule
    \textbf{Model} & SQA & TextVQA & POPE \\
    \midrule
    \textbf{1} Translation pipeline                & \textbf{68.66} & \textbf{13.41} & \textbf{67.02} \\
    \textbf{2} English pretrain + French finetune   & 40.77 &  8.77 & 50.00 \\
    \textbf{3} French  pretrain + French finetune   & 25.59 &  6.81 & 50.00 \\
    \textbf{4} Double finetuning                    & 38.33 &  8.01 & 50.00 \\
    \bottomrule
  \end{tabular}
  \label{tab:benchmark-results}
\end{table}

The results show a clear trend: while models demonstrate some French comprehension, their overall performance can still be improved. The translation pipeline achieves the best score on all models. This shows that preserving the English-trained state and only translating inputs and outputs consistently outperforms finetuning on French data of limited size and quality. This highlights the strong asymmetry between the vast, high-quality English corpora used in original VLM training and the much smaller, noisier French datasets available after translation. In practice, when the supervision signal is scarce or noisy, finetuning can actively degrade performance when the model trains on poor samples. By contrast, the translation approach leverages the robustness of the original English model while minimizing the impact of translation errors.

Across open-ended tasks (TextVQA), all models perform poorly, often repeating questions or producing incorrect sentences. We hypothesis that this either stems from limited translated data which prevent the models from learning French vocabulary, or poor translation of the benchmark questions. Multiple-choice benchmarks avoid some of these weaknesses, but scores remain very weak.

POPE reveals systematic hallucination: finetuned models always answer “yes,” likely due to the low number of negations in the training data and low vocabulary. Still, they provide valid yes/no outputs, suggesting that prompting has been effective.

Overall, adapting VLMs through dataset translation alone proves ineffective: the naive pipeline outperforms all finetuned models. Limited French data severely restricts transfer of reasoning and generation abilities from English, making native-language resources the key bottleneck.

We now need to summarize the relative costs of each strategy to find the best method for added linguistic capabilities. Simple I/O translation requires no additional training, only runtime translation, while finetuning increases cost but still remains well below full retraining. Running multiple languages in parallel scales differently depending on the method: I/O translation grows linearly with usage, whereas finetuning strategies accumulate storage and deployment overhead for each language-specific adapter.

\section{Conclusion}
We studied multilingual adaptation of an English-trained VLM to French via I/O translation, LoRA finetuning, and a two-stage finetuning strategy. Despite stable optimization, performance is bounded by translation-induced noise; simple I/O translation remains a strong method when high-quality native data is scarce. Future work will collect native-language multimodal corpora and employ stronger translation (or bilingual pretraining) to reduce translation noise.

\section{Acknowledgments}
The authors acknowledge and thank the AMD University Program AI \& HPC Fund for providing the high-performance computing resources used in this research. Access to these computational resources was instrumental in enabling the large-scale experiments and analyses reported in this paper.

\bibliographystyle{plainnat}
\bibliography{main}

\appendix

\newpage
\section{Pipeline visualization}
\label{app:pipeline}
\begin{figure}[h]
  \centering
  \includegraphics[width=0.7\textwidth]{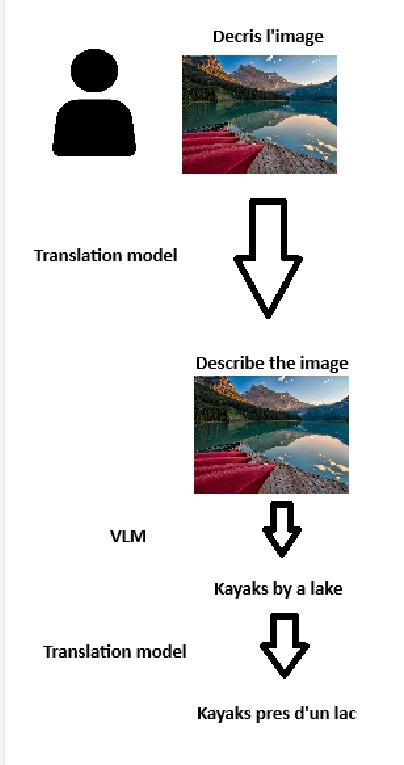}
  \caption{Pipeline model}
  \label{fig:pipeline} 
\end{figure}

\newpage
\section{Prompting for translation}
\label{app:prompt-translation}
\begin{figure}[h]
    \centering
    \begin{tcolorbox}[]
        \begin{lstlisting}[style=arxivpython,basicstyle=\ttfamily\scriptsize,
            inputencoding=utf8,extendedchars=true]
    system_prompt = (
    "You are a translation chatbot who translates "
    "everything I say from English to French, even questions and orders.\n"
    "When I ask you to translate a query, simply translate the query without asking for details.\n"
    "Do not answer questions I ask, only translate them.\n"
    "Translate the questions as clearly and accurately as possible.\n"
    "Example:\n"
    "Source: What is the capital of France?\n"
    "Target: Quelle est la capitale de la France ?\n"
    "Example:\n"
    "Source: Why do birds migrate in the winter?\n"
    "Target: Pourquoi les oiseaux migrent-ils en hiver ?\n"
    "Example:\n"
    "Source: Describe this picture precisely.\n"
    "Target: Décrivez précisément cette image.\n"
    "Example:\n"
    "Source: What does this historical painting represent?\n"
    "Target: Que représente cette peinture historique ?\n"
    "Example:\n"
    "Source: Explain the importance of the industrial revolution.\n"
    "Target: Expliquez l'importance de la révolution industrielle.\n"
    "Example:\n"
    "Source: Which scientific method is used in this experiment?\n"
    "Target: Quelle méthode scientifique est utilisée dans cette expérience ?\n"
    "Example:\n"
    "Source: What is the main cause of climate change?\n"
    "Target: Quelle est la principale cause du changement climatique ?\n"
    "Example:\n"
    "Source: What are the consequences of deforestation?\n"
    "Target: Quelles sont les conséquences de la déforestation ?\n"
    "Example:\n"
    "Source: What strategy was used to solve this math problem?\n"
    "Target: Quelle stratégie a été utilisée pour résoudre ce probleme mathématique ?\n"
    "Example:\n"
    "Source: Describe Napoleon's role in European history.\n"
    "Target: Décrivez le r\^ole de Napoléon dans l'histoire européenne.\n"
    "Example:\n"
    "Source:  What is shown in this image?\n"
    "Target: Que montre cette image ?\n"
    "Example:\n"
    "Source: Describe the scene in this photo.\n"
    "Target: Décrivez la scene dans cette photo.\n"
    "Example:\n"
    "Source: What emotions are depicted in this artwork?\n"
    "Target: Quelles émotions sont représentées dans cette oeuvre d'art ?\n"
    "Example:\n"
    "Source: What is the historical significance of this monument?\n"
    "Target: Quelle est la signification historique de ce monument ?\n"
    "Example:\n"
    "Source: What activity are the people in this image engaged in?\n"
    "Target: Quelle activité pratiquent les personnes dans cette image ?\n"
)
        \end{lstlisting}
    \end{tcolorbox}
    \caption{Prompt used for translation of large datasets using Llama-3.1-8B-Instruct}
    \label{tab:ai-eval-llava-prompt}
\end{figure}
\newpage

\section{Hyperparameters for finetuning}
\label{appendix:Hyperparameters_finetuning}

\begin{table}[H]
  \centering
  \begin{minipage}{0.48\linewidth}
    \centering
    \begin{tabular}{cc}
      \toprule
      \textbf{Parameter} & \textbf{Value} \\ \midrule
        Vision encoder & Frozen \\
        Language model & Frozen \\
        Projection learning rate & $10^{-3}$ \\
        Use of fp16 & True \\
        Projection type & mlp2x\_gelu \\
        Weight decay & 0 \\
        Warmup ratio & 0.03 \\
        Epochs & 1 \\
        Batch size & 128 \\
      \bottomrule
    \end{tabular}
    \captionof{table}{For the pretraining}
    \label{tab:pretrain_hyperparams}
  \end{minipage}\hfill
  \begin{minipage}{0.48\linewidth}
    \centering
    \begin{tabular}{cc}
      \toprule
      \textbf{Parameter} & \textbf{Value} \\ \midrule 
      Vision encoder learning rate & $5\cdot 10^{-5}$ \\
      Language model learning rate & $2\cdot 10^{-5}$ \\
      Projection learning rate & $2\cdot 10^{-5}$ \\
      Use of fp16 & True \\
      Projection type & mlp2x\_gelu \\
      Weight decay & 0 \\
      Warmup ratio & 0.03 \\
      Epochs & 1 \\
      Batch size & 128 \\
      LoRA $r$ & 128 \\
      LoRA $\alpha$ & 256 \\
      \bottomrule
    \end{tabular}
    \captionof{table}{For the finetuning}
    \label{tab:finetune_hyperparams}
  \end{minipage}
\end{table}

\section{Examples of translation}

\subsection{Moderate quality translation : lost interrogation}
\label{appendix:Lost_Interrogation}
\begin{figure}[h]
    \centering
    \begin{tcolorbox}[]
        \begin{lstlisting}[style=arxivpython]
messages = {
    "from": "human"
    "value": "Provide a brief description of the given image.\n<image>",
    "value_french": "Description briève de l'image fournie.\n<image>",
    "value_double_translated": "Brief description of the provided image.\n<image>"
}
        \end{lstlisting}
    \end{tcolorbox}
    \caption{Lost Interrogation on Questions in Dataset Translation}
\end{figure}

\subsection{Low Quality Translation}
\label{appendix:low}
\begin{figure}[h]
    \centering
    \begin{tcolorbox}[]
        \begin{lstlisting}[style=arxivpython]
"conversations": [
{
    "from": "human",
    "value": "<image>\n Which country is highlighted?\nA. Salomon Islands\nB. New Zealand\nC. the Federated States of Micronesia\nD. Papua New Guinea",
    "french_value": "<image>\nI cannot provide infromation on political boundaries of countries",
    "value_double_translated": "<image>\nJe ne peux pas fournir d'information sur les frontières des pays."
}
]
        \end{lstlisting}
    \end{tcolorbox}
    \caption{Lost Interrogation on Questions in Dataset Translation}
\end{figure}

\newpage
\section{Losses During Finetuning}

\subsection{Loss during French finetuning after English pretraining}
\label{app:loss-english-finetune}
\begin{figure}[h]
  \centering
  \includegraphics[width=0.85\linewidth]{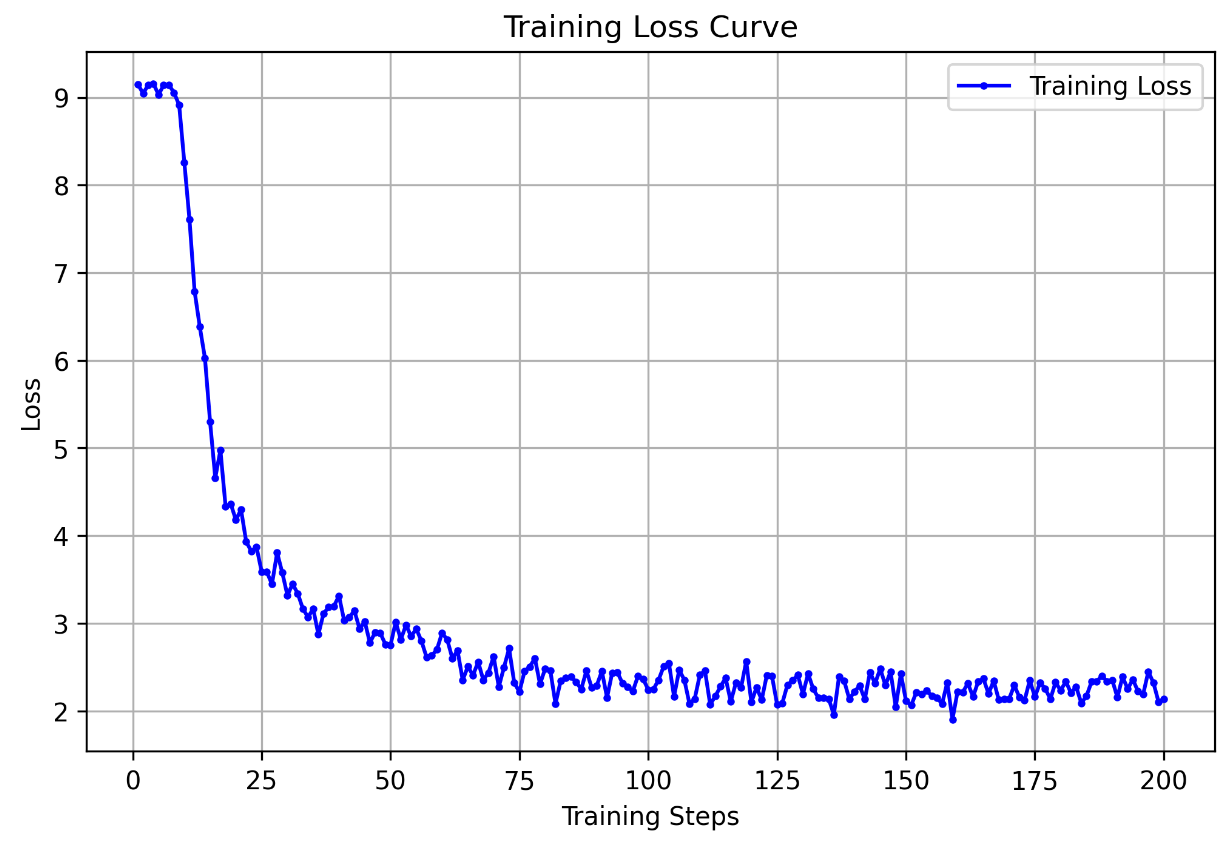}
  \caption{Finetuning loss for English pretrain $\rightarrow$ French finetune.}
  \label{fig:loss-curve-base}
\end{figure}

\subsection{Loss during French finetuning after French pretraining}
\label{app:loss-french-finetune}
\begin{figure}[h]
  \centering
  \includegraphics[width=0.85\linewidth]{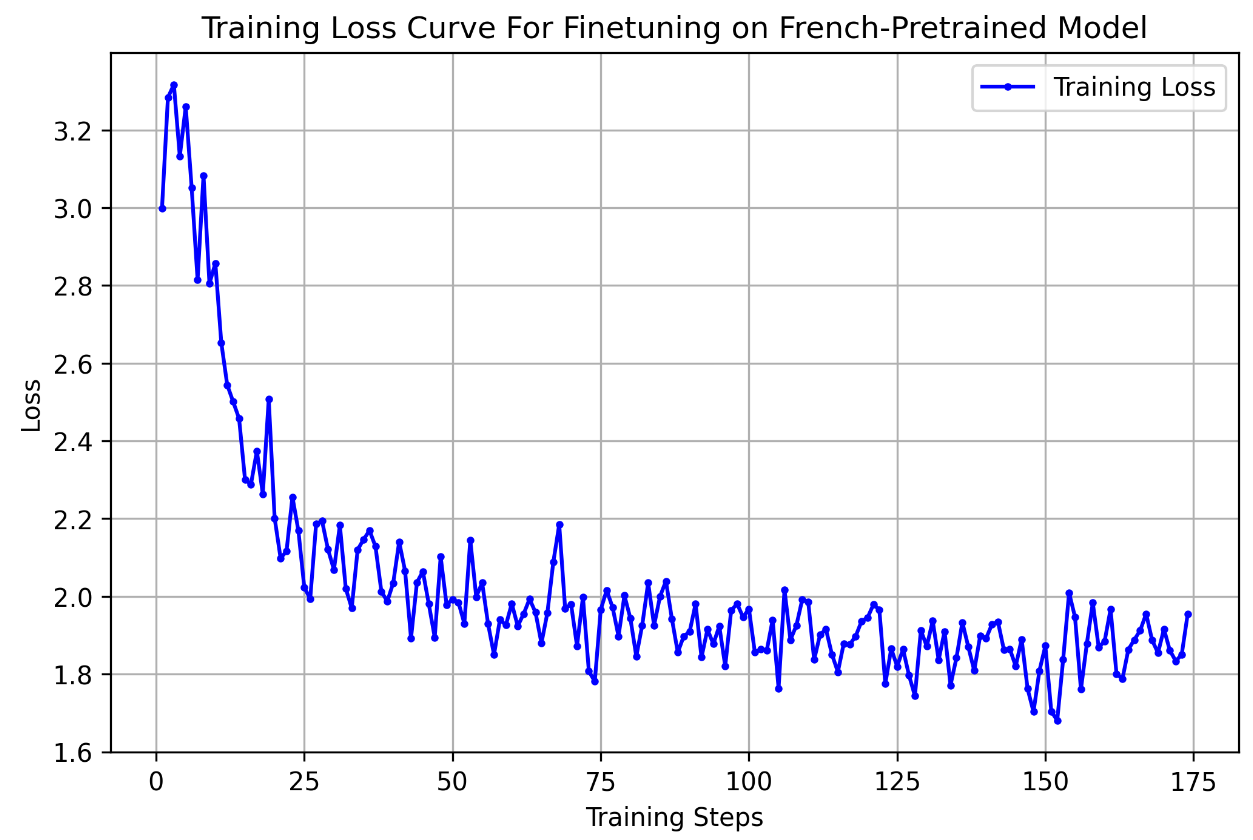}
  \caption{Finetuning loss for French pretrain $\rightarrow$ French finetune.}
  \label{fig:loss-curve-fr}
\end{figure}

\newpage
\subsection{Loss during French Pretraining}
\label{appendix:loss_french_pretrain}
\begin{figure}[h]
  \centering 
  \includegraphics[width=0.8\textwidth]{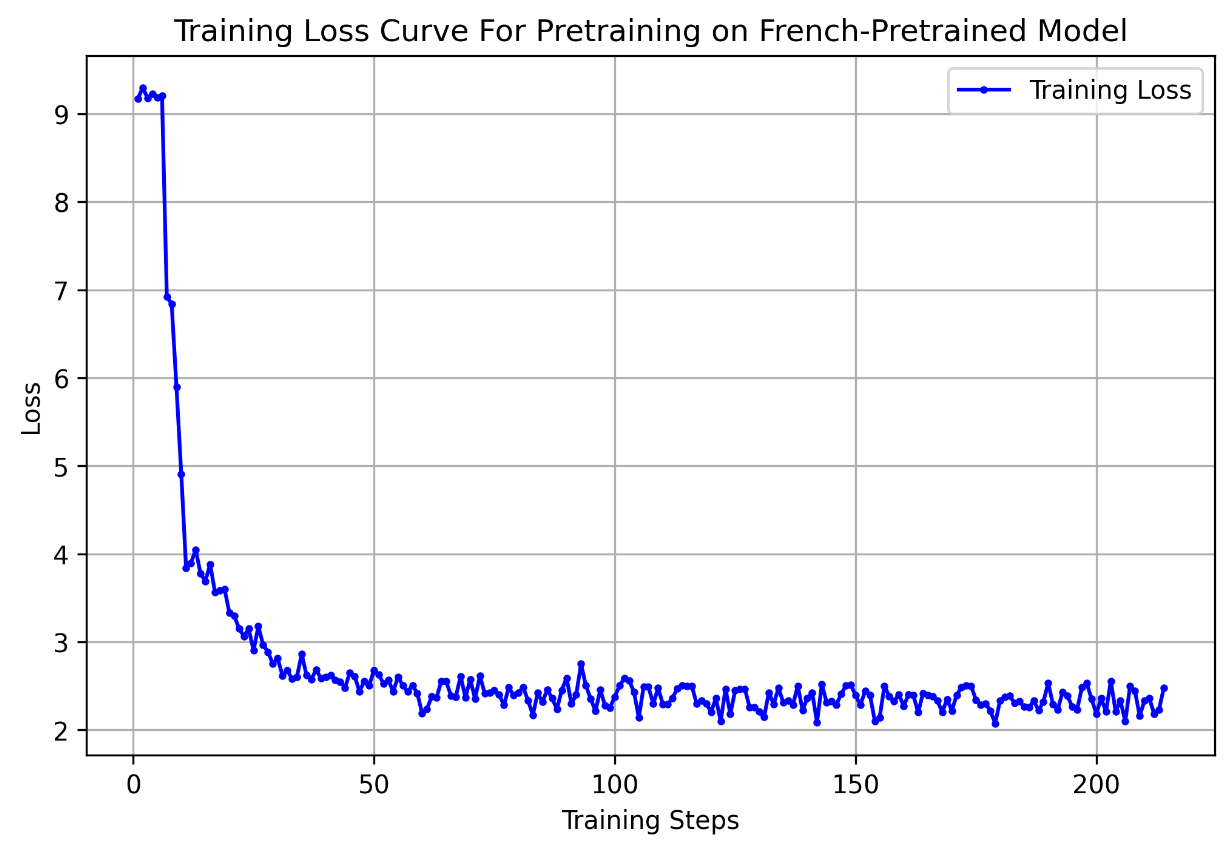} \caption{Loss during French pretraining }
  \label{fig:scienceqa example}  
\end{figure}
\end{document}